\pdfoutput=1
\documentclass[11pt]{article}

\usepackage{acl}

\usepackage{times}
\usepackage{latexsym}
\usepackage{graphicx}
\usepackage{arydshln}
\usepackage{subfigure}
\usepackage{makecell}
\usepackage{colortbl}
\usepackage{pifont}
\usepackage{amssymb}

\usepackage[T1]{fontenc}
\usepackage{xcolor}
\usepackage[utf8]{inputenc}

\usepackage{microtype}

\usepackage{array}
\usepackage{pifont}
\usepackage{tabularx}
\usepackage{adjustbox}
\usepackage{multirow}
\usepackage{enumitem}
\usepackage{xspace}
\usepackage{tcolorbox}
\usepackage{xparse}
\usepackage{soul}
\usepackage{booktabs,amsfonts,dcolumn}
\usepackage{hyperref}
\usepackage{url}
\usepackage{amsmath,amsthm,amsfonts,amssymb,bm,stmaryrd}
\usepackage[noorphans,vskip=0.75ex,leftmargin=2ex]{quoting}
\usepackage{breqn}

\definecolor{OliveGreen}{rgb}{0,0.4,0}

\usepackage{cleveref}

\crefname{section}{§}{§§}
\Crefname{section}{§}{§§}

\title{Structured Pruning Learns Compact and Accurate Models}

\author{Mengzhou Xia \quad Zexuan Zhong \quad Danqi Chen \\
Princeton University \\
\ttt{\{mengzhou,zzhong,danqic\}@cs.princeton.edu}\\
}

\newcommand\ti[1]{\textit{#1}}

\newcommand\tf[1]{\textbf{#1}}
\newcommand\ttt[1]{\texttt{#1}}
\newcommand\mf[1]{\mathbf{#1}}

\newcommand{\tableindent}{~~}

\DeclareMathOperator*{\argmin}{arg\,min}
\renewcommand{\paragraph}[1]{\vspace{0.2cm}\noindent\textbf{#1}}

\newcommand{\ours}{CoFi}

\newcommand{\mha}{MHA}
\newcommand{\ffn}{FFN}
\newcommand{\ccmark}{\ding{51}}%
\newcommand{\xxmark}{\ding{55}}%

\usepackage{xcolor}
\usepackage{xparse}

\usepackage{soul}

\definecolor{color_m}{RGB}{72,117,170}
\definecolor{color_f}{RGB}{201,89,72}
\definecolor{color_c}{RGB}{230,230,230}
\definecolor{color_e}{RGB}{100,155,74}
\definecolor{ccon}{HTML}{fee9d4}
\definecolor{cood}{HTML}{d8f0d3}
\definecolor{cid}{HTML}{dae8f5}
\definecolor{gg}{HTML}{e2f0cb}

\newcommand{\CC}{\cellcolor{gray!15}}

\ExplSyntaxOn

\NewDocumentCommand{\fs}{m}
 {
  \tl_map_function:nN { #1 } \firth_fs:n
 }

\cs_new_protected:Nn \firth_fs:n
 {
  \str_case:nn { #1 }
  {
    {m}{\__firth_fs_box:nn { color_m } { M }}
    {f}{\__firth_fs_box:nn { color_f } { F }}
    {n}{\__firth_fs_box:nn { color_c } { ~ }}
    {g}{\__firth_fs_box:nn { color_c } { ~ }}
    {E}{\__firth_fs_box:nn { color_e } { E:d }}
    {M}{\__firth_fs_box:nn { color_m } { M:v$\times$h }}
    {F}{\__firth_fs_box:nn { color_f } { F:m }}
  }
 }
\cs_new_protected:Nn \__firth_fs_box:nn
 { %{\setlength{\fboxsep}{0pt}\colorbox{<color>}{<text>}}
  {\setlength{\fboxsep}{1.3pt}\colorbox{#1}
  {
    \color{white}
    \use:c { check@mathfonts }
    \fontsize { \use:c{ sf@size } } { 0 } \selectfont
     \texttt{\textbf{#2}} \vphantom{f|} 
  }}% \thinspace
 }

\ExplSyntaxOff
\newcommand{\fst}[1]{\Large{\fs{#1}}}

\begin{document}

\NewDocumentCommand\cofi{}{\includegraphics[width=0.35cm]{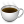}}

\maketitle
\begin{abstract}

The growing size of neural language models has led to increased attention in model compression.
The two predominant approaches are \ti{pruning}, which gradually removes weights from a pre-trained model, and \ti{distillation}, which trains a smaller compact model to match a larger one.
Pruning methods can significantly reduce the model size but hardly achieve large speedups as distillation. However, distillation methods require large amounts of unlabeled data and are expensive to train.
In this work, we propose a task-specific structured pruning method {{\ours}}\footnote{CoFi is pronounced as \cofi.} (\tf{Co}arse- and \tf{Fi}ne-grained Pruning), which delivers highly parallelizable subnetworks and matches the distillation methods in both accuracy and latency, without resorting to any unlabeled data. Our key insight is to jointly prune coarse-grained (e.g., layers) and fine-grained (e.g., heads and hidden units) modules, which controls the pruning decision of each parameter with masks of different granularity. We also devise a layerwise distillation strategy to transfer knowledge from unpruned to pruned models during optimization. Our experiments on GLUE and SQuAD datasets show that {{\ours}} yields models with over 10$\times$ speedups with a small accuracy drop, showing its effectiveness and efficiency compared to previous pruning and distillation approaches.\footnote{Our code and models are publicly available at \url{https://github.com/princeton-nlp/CoFiPruning}.}

\end{abstract}

%!TEX root = ../main.tex

\section{Introduction}

%!TEX root = ../main.tex

\begin{table}[!t]
    \centering
    \resizebox{0.95\columnwidth}{!}{%
    \begin{tabular}{lccrrr}
    \toprule
   &  \ti{U} &  \ti{T} & $\nearrow$ & {Params}  & {MNLI}\\
    \midrule
    BERT$_{\text{base}}$ (teacher) & \xxmark & \xxmark & $1.0 \times$ & $85$M & $84.8$ \\
    \midrule
    \tf{Distillation} \\
    \tableindent DistillBERT$_6$ & \ccmark & \xxmark  & ${2.0\times}$  & $43$M & $82.2$ \\
    % \tableindent DistillBERT$_6$ & ${2.0\times}$  & $43$M & $82.2$
    \tableindent TinyBERT$_6$ & \ccmark & \ccmark &   ${2.0\times}$ & $43$M   & $84.0$ \\
    % \tableindent TinyBERT$_6$ w/ DA&  ${2.0\times}$ & $43$M   & $84.6$ \\
    \tableindent MobileBERT${^{\ddagger}}$ & \ccmark & \xxmark  & $2.3\times$ & $20$M  & $83.9$ \\
    \tableindent DynaBERT & \xxmark & \ccmark   & $6.3\times$ & $11$M & $76.3$ \\
    \tableindent  \cellcolor{gg}{AutoTinyBERT${^{\ddagger}}$} & \cellcolor{gg}\ccmark & \cellcolor{gg}\ccmark  & \cellcolor{gg}{$9.1\times$}& \cellcolor{gg}{$3.3$M} &\cellcolor{gg}{$78.2$} \\
    \tableindent \cellcolor{gg}{TinyBERT$_4$} & \cellcolor{gg}\ccmark & \cellcolor{gg}\ccmark & \cellcolor{gg}{$11.4\times$} & \cellcolor{gg}{$4.7$M} & \cellcolor{gg}{$78.8$} \\
    % \tableindent TinyBERT$_4$  & ${11.4\times}$& $4.7$M & $80.5$ \\
    % \tableindent TinyBERT$_4$ w/ DA  & ${11.4\times}$& $4.7$M & $82.8$ \\

    \midrule
    \tf{Pruning} \\
    \tableindent Movement Pruning & \xxmark & \ccmark &  $1.0\times$ & $9$M & $81.2$ \\
    % \tableindent FLOP & $2.0\times$ & $34$M & $82.2$  \\
    \tableindent Block Pruning &\xxmark & \ccmark & $2.7\times$ & $25$M& $83.7$\\
    % \tableindent BlockPruning & $11$M & $5.0\times$ & $81.2$ \\

    \tableindent {\ours} Pruning (ours) & \xxmark & \ccmark & $2.7\times$ & $26$M & $84.9$\\
    \tableindent \cellcolor{gg}{\ours} Pruning (ours) & \cellcolor{gg}\xxmark & \cellcolor{gg}\ccmark & \cellcolor{gg}{$12.1\times$} & \cellcolor{gg}{$4.4$M} & \cellcolor{gg}{$80.6$} \\
    \bottomrule
    \end{tabular}
    }
    \caption{A comparison of state-of-the-art distillation and pruning methods. \ti{U} and \ti{T} denote whether \ti{U}nlabeled and \ti{T}ask-specific are used for distillation or pruning. The inference speedups ($\nearrow$) are reported against a BERT$_{\text{base}}$ model and we evaluate all the models on an NVIDIA V100 GPU (\S\ref{sec:exp_setup}). The models labeled as $^\ddagger$ use a different teacher model and are not a direct comparison. \colorbox{gg}{Models} are one order of magnitude faster.\protect\footnotemark}
    \label{tab:pruning_vs_distillation}
\end{table}

Pre-trained language models ~\cite[][\emph{inter alia}]{devlin2019bert,liu2019roberta,raffel2020exploring} have become the mainstay in natural language processing. These models have high costs in terms of storage, memory, and computation time and it has motivated a large body of work on model compression to make them smaller and faster to use in real-world applications~\cite{ganesh2021compressing}.

The two predominant approaches to model compression are pruning and distillation (Table~\ref{tab:pruning_vs_distillation}).

Pruning methods search for an accurate subnetwork in a larger pre-trained model.
Recent work has investigated how to structurally prune Transformer networks~\cite{vaswani2017attention}, from removing entire layers~\cite{fan2020reducing,sajjad2020poor}, to pruning heads~\cite{michel2019sixteen,voita2019analyzing}, intermediate dimensions~\cite{mccarley2019structured,wang2020structured} and blocks in weight matrices~\cite{lagunas2021block}. The trend of structured pruning leans towards removing fine-grained units to allow for flexible final structures. However,  thus far, pruned models rarely achieve large speedups~(2-3$\times$ improvement at most).

\footnotetext{Following previous work, we exclude embedding matrices in calculating the number of parameters. We exclude task-specific data augmentation for a fair comparison. More results with data augmentation can be found in \autoref{tab:da}. }

\begin{figure*}[t]
\centering
\includegraphics[width=0.85\linewidth]{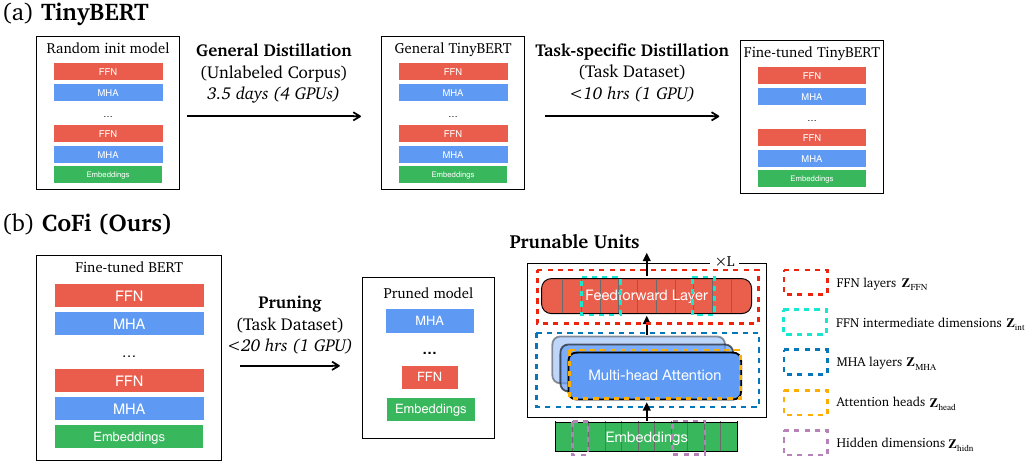}
% \vspace{-0.5em}
\caption{
Comparison of (a)  TinyBERT~\cite{jiao2020tinybert} and (b) our pruning approach {\ours}. TinyBERT trains a randomly-initialized network through two-step distillation: (1) general distillation on a large unlabeled corpus, which takes 3.5 days to finish on 4 GPUs, and (2) task-specific distillation on the task dataset. {\ours} directly prunes the fine-tuned BERT model and jointly learns five types of mask variables (i.e., $\mathbf{z}_{\mathrm{FFN}}, \mathbf{z}_{\mathrm{int}}, \mathbf{z}_{\mathrm{MHA}}, \mathbf{z}_{\mathrm{head}}, \mathbf{z}_{\mathrm{hidn}}$) to prune different types of units (\S\ref{sec:mixed_pruning}).
{\ours} takes at most 20 hours to finish on 1 GPU on all the GLUE datasets (smaller datasets need $<3$ hours).\protect\footnotemark}
\label{fig:teaser}
\end{figure*}

By contrast, distillation methods usually first specify a fixed model architecture and perform a \ti{general distillation} step on an unlabeled corpus, before further fine-tuning or distillation on task-specific data~\cite{sanh2019distilbert,turc2019well, sun2019patient, jiao2020tinybert}.
Well-designed student architectures achieve compelling speedup-performance tradeoffs, yet distillation to these randomly-initialized student networks on large unlabeled data is prohibitively slow.\footnote{There are exceptions like DistillBERT~\cite{sanh2020movement}, which initializes the student from the teacher by taking one layer out of two, yet it is unclear how to generalize this initialization scheme to other compact structures.} For instance, TinyBERT~\cite{jiao2020tinybert} is first trained on 2,500M tokens for 3 epochs, which requires training 3.5 days on 4 GPUs (Figure~\ref{fig:teaser}).\footnote{See training time measurement details in \autoref{sec:train_time}.}

In this work, we propose a task-specific, structured pruning approach called {\ours} (\textbf{Co}arse and \textbf{Fi}ne-grained Pruning) and show that structured pruning can achieve highly compact subnetworks and obtain large speedups and competitive accuracy as distillation approaches, while requiring much less computation.
Our key insight is to jointly prune coarse-grained units (e.g., self-attention or feed-forward layers) and fine-grained units (e.g., heads, hidden dimensions) simultaneously. Different from existing works, our approach controls the pruning decision of every single parameter by multiple masks of different granularity. This is the key to large compression, as it allows the greatest flexibility of pruned structures and eases the optimization compared to only pruning small units.

It is known that pruning with a distillation objective can substantially improve performance~\cite{sanh2020movement,lagunas2021block}.
Unlike a fixed student architecture, pruned structures are unknown prior to training and it is challenging to distill between intermediate layers of the unpruned and pruned models~\cite{jiao2020tinybert}. Hence, we propose a layerwise distillation method,  which dynamically learns the layer mapping between the two structures. We show that this strategy can better lead to performance gains beyond simple prediction-layer distillation.

Our experiments show that {\ours} delivers more accurate models at all levels of speedups and model sizes on the GLUE~\cite{wang2019glue} and SQuAD v1.1~\cite{rajpurkar2016squad} datasets, compared to strong pruning and distillation baselines. Concretely, it achieves over 10$\times$ speedups and a 95\% sparsity across all the datasets while preserving more than 90\% of accuracy.
% The performance is comparable to TinyBERT~\cite{jiao2020tinybert} and AutoTinyBERT~\cite{yin2021autotinybert}.
Our results suggest that task-specific structured pruning is an appealing solution in practice, yielding smaller and faster models without requiring additional unlabeled data for general distillation.

%!TEX root = ../main.tex

\section{Background}

\subsection{Transformers}
\label{sec:transformers}
A Transformer network~\cite{vaswani2017attention} is composed of $L$ blocks and each block consists of a multi-head self-attention (\mha) layer, and a feed-forward (\ffn) layer.
An {\mha} layer with $N_h$ heads takes an input $X$ and outputs: %  \in \mathbb{R}^{\mathrm{len} \times d}
\begin{equation}
\resizebox{\hsize}{!}{$
    \mathrm{\mha} (X) =
    \sum_{i=1}^{N_h} \mathrm{Att}(W_Q^{(i)}, W_K^{(i)}, W_V^{(i)}, W_O^{(i)}, X), \nonumber
$
}
\end{equation}
where $W^{(i)}_Q, W^{(i)}_K, W^{(i)}_V, W^{(i)}_O \in \mathbb{R}^{d \times d_h}$ denote the query, key, value and output matrices respectively and $\mathrm{Att}(\cdot)$ is an attention function. Here $d$ denotes the hidden size (e.g., 768) and $d_h = d / N_h$ denotes the output dimension of each head (e.g., 64).

Next comes a feed-forward layer, which consists of an up-projection and a down-projection layer, parameterized by ${W_U} \in \mathbb{R}^{d \times d_f}$ and ${W_D} \in \mathbb{R}^{d_f \times d}$:
\begin{equation}
    \mathrm{\ffn}(X) = \mathrm{gelu}(XW_U) \cdot W_D \nonumber.
\end{equation}
Typically, $d_f = 4d$. There is also a residual connection and a layer normalization operation after each {\mha} and {\ffn} layer.

{\mha}s, {\ffn}s account for $1/3$ and $2/3$ of the model parameters in Transformers (embeddings excluded). According to \citet{ganesh2021compressing}, both {\mha}s and {\ffn}s take similar time on GPUs while {\ffn}s become the bottleneck on CPUs.

\subsection{Distillation}
\label{subsec:dis}
Knowledge distillation~\cite{hinton2015distilling} is a model compression approach that transfers knowledge from a larger teacher model to a smaller student model. \ti{General distillation}~\cite{sanh2019distilbert, sun2020mobilebert, wang2020minilm} and \ti{task-specific distillation} \cite{sun2019patient} exploit unlabeled data and task-specific data respectively for knowledge transfer. A combination of the two leads to increased performance \cite{jiao2020tinybert}. General distillation or pre-training the student network on unlabeled corpus is essential for retaining performance while being computationally expensive~\cite{turc2019well, jiao2020tinybert}.

Different distillation objectives have been also explored. Besides standard distillation from the prediction layer~\cite{hinton2015distilling}, transferring knowledge layer-by-layer from representations \cite{jiao2020tinybert, sun2020mobilebert} and multi-head attention matrices~\cite{wang2020minilm, jiao2020tinybert, sun2020mobilebert} lead to significant improvements. Most distillation approaches assume a fixed student structure prior to training. \citet{hou2020dynabert} attempt to distill to a dynamic structure with specified widths and heights. \citet{yin2021autotinybert} adopt a one-shot Neural Architecture Search solution to search architectures of student networks.

\subsection{Pruning}

\footnotetext{{\ours} requires slightly longer training time compared to the task-specific distillation of TinyBERT, as {\ours} searches model structures and learns parameters simultaneously.}

\label{subsec:bgpruning}

Pruning gradually removes redundant parameters from a teacher model, mostly producing task-specific models. Previous works focus on pruning different components in Transformer models, from coarse-grained to fine-grained units.

\paragraph{Layer pruning} \citet{fan2020reducing} and \citet{sajjad2020poor} explore strategies to drop entire Transformer blocks (a pair of {\mha} and {\ffn} layer) from a pre-trained model. Empirical evidence suggests that 50\% of layers can be dropped without a big accuracy drop, resulting in a $2\times$ speedup.

\paragraph{Head pruning}
\citet{voita2019analyzing,michel2019sixteen} show that only a small subset of heads are important and the majority can be pruned. We follow these works to mask heads by introducing variables $\mathbf{z}^{(i)}_{\mathrm{head}} \in \{0, 1\}$ to multi-head attention:
\begin{equation}
% \begin{split}
\resizebox{\hsize}{!}{$
    \mathrm{MHA} (X) = \\
    \sum_{i=1}^{N_h} \mathbf{z}_{\mathrm{head}}^{(i)} \mathrm{Att}(W_Q^{(i)}, W_K^{(i)}, W_V^{(i)}, W_O^{(i)}, X). \nonumber
$}
% \end{split}
\end{equation}
Only removing heads does not lead to large latency improvement---\citet{li2021differentiable} demonstrate a 1.4$\times$ speedup with only one remaining head per layer.

\paragraph{{\ffn} pruning} The other major part---feed-forward layers ({\ffn}s)---are also known to be overparameterized. Strategies to prune an {\ffn} layer for an inference speedup include pruning an entire {\ffn}  layer~\cite{prasanna2020bert, chen2020earlybert} and at a more fine-grained level, pruning intermediate dimensions~\cite{mccarley2019structured, hou2020dynabert} by introducing $\mathbf{z}_{\mathrm{int}} \in \{0, 1\}^{d_f}$:
\begin{equation}
    \mathrm{\ffn}(X) = \mathrm{gelu}(XW_U) \cdot \mathrm{diag}(\mathbf{z}_{\mathrm{int}}) \cdot W_D. \nonumber
\end{equation}

\paragraph{Block and unstructured pruning} More recently, pruning on a smaller unit, blocks, from {\mha}s and {\ffn}s have been explored~\cite{lagunas2021block}. However, it is hard to optimize models with blocks pruned thus far: \citet{yao2021mlpruning} attempt to optimize block-pruned models with the block sparse MatMul kernel provided by Triton \cite{tillet2019triton}, but the reported results are not competitive. Similarly, \ti{unstructured pruning} aims to remove individual weights and has been extensively studied in the literature~\cite{chen2020lottery, huang2021sparse}.  Though the sparsity reaches up to 97\% \cite{sanh2020movement}, it is hard to obtain inference speedups on the current hardware.

\paragraph{Combination with distillation} Pruning is commonly combined with a prediction-layer distillation objective~\cite{sanh2020movement, lagunas2021block}. Yet it is not clear how to apply layerwise distillation strategies as the pruned student model's architecture evolves during training.

%!TEX root = ../main.tex

\section{Method}

We propose a structured pruning approach {\ours}, which jointly prunes \tf{Co}arse-grained and \tf{Fi}ne-grained units~(\S\ref{sec:mixed_pruning}) with a layerwise distillation objective transferring knowledge from unpruned to pruned models~(\S\ref{sec:distillation}). A combination of the two leads to highly compressed models with large inference speedups.

\subsection{Coarse- and Fine-Grained Pruning}
\label{sec:mixed_pruning}

Recent trends in structured pruning move towards pruning smaller units for model flexibility. Pruning fine-grained units naturally entails pruning coarse-grained units---for example, pruning $N_h$ (e.g., 12) heads is equivalent to pruning one entire MHA layer. However, we observe that this rarely happens in practice and poses difficulty to optimization especially at a high sparsity regime.

To remedy the problem, we present a simple solution: we allow pruning {\mha} and {\ffn} layers explicitly along with fine-grained units  (as shown in \S\ref{subsec:bgpruning}) by introducing two additional masks ${z}_{\mathrm{MHA}}$ and ${z_{\mathrm{FFN}}}$ for each layer. Now the multi-head self-attention and feed-forward layer become:

\begin{align*}
        &\mathrm{MHA} (X) = {{z}_{\mathrm{MHA}}} \cdot \sum_{i=1}^{N_h}  (\mathbf{z}_{\mathrm{head}}^{(i)} \cdot  \\
    &~~~~~~~~~~~~~~~~~~~~~\mathrm{Att}(W_Q^{(i)}, W_K^{(i)}, W_V^{(i)}, W_O^{(i)}, X)), \nonumber \\
    &\mathrm{FFN} (X) = {{z}_{\mathrm{FFN}}} \cdot
      \mathrm{gelu}(X W_U) \cdot \mathrm{diag}(\mathbf{z}_{\mathrm{int}}) \cdot W_D.
\end{align*}

With these layer masks, we explicitly prune an entire layer, instead of pruning all the heads in one {\mha} layer (or all the intermediate dimensions in one {\ffn} layer).  Different from the layer dropping strategies in \newcite{fan2020reducing, sajjad2020poor}, we drop {\mha} and {\ffn} layers separately, instead of pruning them as a whole.

 % the pruning decisions for layers are directly learned, instead of indirectly accumulated from pruning heads or intermediate dimensions.

Furthermore, we also consider pruning the output dimensions of $\mathrm{MHA}(X)$ and $\mathrm{FFN}(X)$, referred to as `hidden dimensions' in this paper, to allow for more flexibility in the final model structure. We define a set of masks $\mathbf{z}_{\mathrm{hidn}} \in \{0, 1\}^{d}$, shared across layers because each dimension in a hidden representation is connected to the same dimension in the next layer through a residual connection. These mask variables are applied to all the weight matrices in the model, e.g., $\mathrm{diag}(\mathbf{z}_{\mathrm{hidn}}) W_Q$. Empirically, we find that only a small number of dimensions are pruned (e.g., $768 \rightarrow 760$), but it still helps improve performance significantly (\S\ref{sec:pruning_units}).

{\ours} differs from previous pruning approaches in that multiple mask variables jointly control the pruning decision of one single parameter. For example, a weight in an FFN layer is pruned when the entire FFN layer, or its corresponding intermediate dimension, or the hidden dimension is pruned.  As a comparison, a recent work Block Pruning~\cite{lagunas2021block} adopts a hybrid approach which applies one single pruning strategy on {\mha}s and {\ffn}s separately.

% different from what has been done in previous pruning approaches.
To learn these mask variables, we use $l_0$ regularization modeled with hard concrete distributions following \citet{louizos2018learning}. We also follow \citet{wang2020structured} to replace the vanilla $l_0$ objective with a Lagrangian multiplier to better control the desired sparsity of pruned models.\footnote{We also tried a straight-through estimator as proposed in \citet{sanh2020movement} and found the performance comparable. We choose $l_0$ regularization because it is easier to control the sparsity precisely.} We adapt the sparsity function accordingly to accommodate pruning masks of different granularity:
\begin{align*}
&\scalebox{0.9}{$
    \hat s = \frac{1}{M} \cdot 4 \cdot d_{h} \cdot \sum_i^L \sum_j^{N_h} \sum_k^{d} \mathbf{z}_{\textrm{MHA}}^{(i)} \cdot \mathbf{z}_{\textrm{head}}^{(i, j)} \cdot \mathbf{z}_{\textrm{hidden}}^{(k)}$}
    \\
&~~~\scalebox{0.9}{$+ \frac{1}{M} \cdot 2 \cdot \sum_i^L \sum_j^{d_f} \sum_k^{d} \mathbf{z}_{\textrm{FFN}}^{(i)} \cdot \mathbf{z}_{\textrm{int}}^{(i, j)} \cdot \mathbf{z}_{\textrm{hidden}}^{(k)}, \nonumber$}
\end{align*}
where $\hat s$ is the expected sparsity and M denotes the full model size. All masking variables are learned as real numbers in $[0, 1]$ during training and we map the masking variables below a threshold to $0$ during inference and get a final pruned structure where the threshold is determined by the expected sparsity of each weight matrix (see \autoref{app:optimization} for more details).

\subsection{Distillation to Pruned Models}
\label{sec:distillation}

Previous work has shown that combining distillation with pruning improves performance, where the distillation objective only involves a cross-entropy loss between the pruned student’s and the teacher’s output probability distributions ${\mathbf{p}_s}$ and ${\mathbf{p}_t}$ ~\cite{sanh2020movement, lagunas2021block}:
\begin{equation}
    \mathcal{L}_{\mathtt{pred}} = D_{\mathtt{KL}}(\mathbf{p}_s~\Vert~ \mathbf{p}_t). \nonumber
\end{equation}
In addition to prediction-layer distillation, recent works show great benefits in distillation of intermediate layers~\cite{sun2019patient,jiao2020tinybert}.

In the context of distillation approaches, the architecture of the student model is pre-specified, and it is straightforward to define a layer mapping between the student and teacher model. For example, the 4-layer TinyBERT$_4$ model distills from the $3$, $6$, $9$ and $12$-th layer of a 12-layer teacher model. However, distilling intermediate layers during the pruning process is challenging as the model structure changes throughout training.

We propose a layerwise distillation approach for pruning to best utilize the signals from the teacher model. Instead of pre-defining a fixed layer mapping, we dynamically search a layer mapping between the full teacher model and the pruned student model. Specifically, let $\mathcal{T}$ denote a set of teacher layers that we use to distill knowledge to the student model.
We define a layer mapping function $m(\cdot)$, i.e., $m(i)$ represents the student layer that distills from the teacher layer $i$.
The hidden layer distillation loss is defined as

\begin{gather}
\mathcal{L}_{\mathtt{layer}} =
\sum_{i \in \mathcal{T}} \mathtt{MSE}(W_{\mathtt{layer}}\mathbf{H}_s^{m(i)}, \mathbf{H}_t^{i}), \nonumber
\end{gather}

\noindent where $W_{\mathtt{layer}} \in \mathbb{R}^{d \times d}$ is a linear transformation matrix, initialized as an identity matrix. $\mathbf{H}_s^{m(i)}, \mathbf{H}_t^{i}$ are hidden representations from $m(i)$-th student {\ffn} layer  and $i$-th teacher {\ffn} layer.
The layer mapping function $m(\cdot)$ is dynamically determined during the training process to match a teacher layer to its closest layer in the student model:
\begin{gather}
m(i) = \argmin_{j: \mathbf{z}_{\mathrm{FFN}}^{(j)} > 0} \mathtt{MSE}(W_{\mathtt{layer}}\mathbf{H}_{s}^{j}, \mathbf{H}_t^{i}). \nonumber
\end{gather}
Calculating the distance between two sets of layers is highly parallelizable and introduces a minimal training overhead. To address the issue of layer mismatch, which mostly happens for small-sized datasets, e.g., RTE, MRPC, we add a constraint to only allow matching a teacher layer to a lower student layer than the previously matched student layer. When pruning with larger sized datasets, layer mismatch rarely happens, showing the superiority of dynamic matching---layers between student and teacher models match in a way that benefits the pruning process the most. 

Finally, we combine layer distillation with the prediction-layer distillation:
\begin{equation}
\mathcal{L}_{\mathtt{distil}} = \lambda \mathcal{L}_{\mathtt{pred}} + (1-\lambda) \mathcal{L}_{\mathtt{layer}}, \nonumber
\end{equation}
where $\lambda$ controls the contribution of each loss.
% + \lambda_1 (s - \hat s) + \lambda_2 (s - \hat s)^2

%!TEX root = ../main.tex

\section{Experiments}

\begin{figure*}[t]
  \centering
  \includegraphics[width=\linewidth]{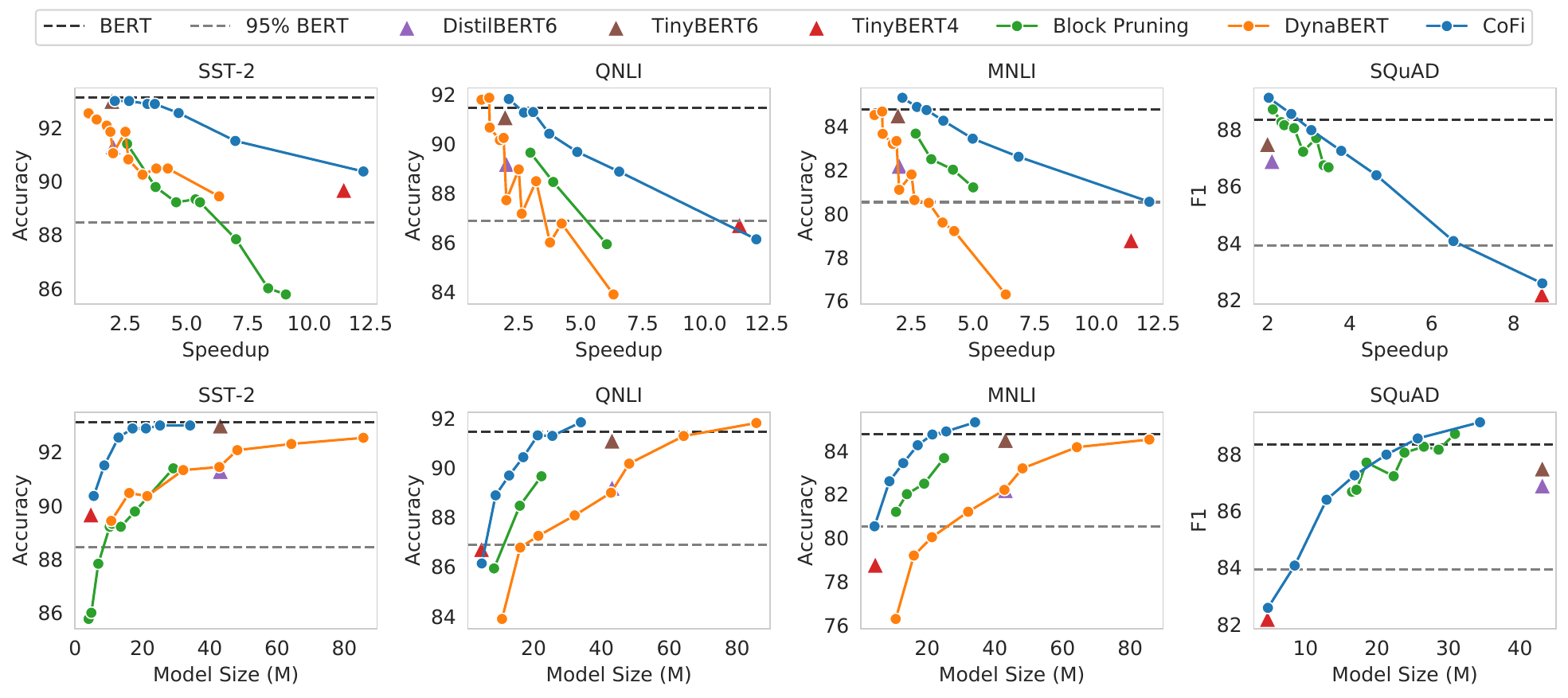}
  \vspace{-2em}
  \caption{Accuracy v.s. speedup (top) or model size (bottom). We compare {\ours} against state-of-the-art distillation and pruning baselines. Note that we exclude embedding size when calculating model size following previous work, as forwarding through the embedding layer has little effect on inference time. }
  \label{fig:main}
\end{figure*}

\subsection{Setup}
\label{sec:exp_setup}

\paragraph{Datasets} We evaluate our approach on eight GLUE tasks~\cite{wang2019glue} and SQuAD v1.1~\cite{rajpurkar2016squad}. GLUE tasks include SST-2~\cite{socher2013recursive_sst-2}, MNLI~\cite{williams2018broad_mnli}, QQP, QNLI, MRPC~\cite{dolan2005automatically}, CoLA~\cite{warstadt2019neural}, STS-B~\cite{cer2017semeval} and RTE (see \autoref{app:data} for dataset sizes and metrics).

\paragraph{Training setup}
In our experiments, \ti{sparsity} is computed as the number of pruned parameters divided by the full model size (embeddings excluded).
Following \newcite{wang2020structured,lagunas2021block}, we first finetune the model with the distillation objective, then we continue training the model with the pruning objective with a scheduler to linearly increase the sparsity to the target value. We finetune the pruned model until convergence (see \autoref{app:hyperparameters} for more training details).

We train models with target sparsities of $\{60\%, 70\%, 75\%, 80\%, 85\%, 90\%, 95\%\}$ on each dataset. For all the experiments, we start from the  BERT$_{\text{base}}$ model\footnote{We also run experiments on RoBERTa models~\cite{liu2019roberta}. Please refer to \autoref{app:roberta} for details.} and freeze embedding weights following \citet{sanh2020movement}.  We report results on development sets of all datasets.

%!TEX root = ../main.tex

% \begin{table}[t]
% \resizebox{1.0\columnwidth}{!}{%
% \begin{tabular}{lcccccccccc} \toprule
% Task & GLUE-L & GLUE-S & SQuAD & Train Time \\\midrule
% BERT$_{\text{base}}$ & $90.2$ &	$76.2$ &	$88.4$ &	- \\ \midrule
% TinyBERT$_4$ w/o GD & $84.4$ &	$37.7$ &	- &	$<10$ \\
% TinyBERT$_4$ & $86.3$ &	$65.5$ &	$82.1$ &	$\sim 350$  \\ \midrule
% {\ours} Pruning (ours)  & $\bf{86.9}$ &	$\bf{66.5}$ &	$\bf{82.6}$ & 	$<20$\\ \bottomrule
% \end{tabular}}
% \vspace{-0.5em}
% \caption{}
% \label{tab:tinybert2}
% \end{table}

\begin{table*}[t]
\resizebox{2.0\columnwidth}{!}{%
\begin{tabular}{lcccccccccc} \toprule
Task & SST-2 & QNLI & MNLI & QQP & CoLA & RTE & STS-B & MRPC & SQuAD & Train Time \\
     & (67k) & (105k) & (393k) & (364k) & (8.5k) & (2.5k) & (7k) & (3.7k) & (88k) \\ \midrule
BERT$_{\text{base}}$ (teacher) & $93.1$ & $91.5$ & $84.8$ & $91.2$ & $61.2$ & $70.0$ & $88.7$ & $85.0$ & $88.4$ & - \\
\midrule
TinyBERT$_4$ w/o GD & $87.7$     & $81.8$    & $78.7$    & $89.5$   & $16.6$    & $47.3$   & $17.8$       & $68.9$    & -       & $\le 10$\\
TinyBERT$_4$ & $89.7$     & $\tf{86.7}$    & $78.8$    & $90.0$     & $32.5$    & $63.2$   & $\tf{85.0}$         & $81.4$    & $82.1$ & $\sim 350$   \\
\rowcolor{gray!15}Speedup  $\nearrow$          & $11.4\times$    & $11.4\times$   & $11.4\times$   & $11.4\times$  & $11.4\times$   & $11.4\times$  & $11.4\times$      & $11.4\times$   & $8.7\times$  & - \\ \midrule
{\ours} Pruning (ours)      & $\textbf{90.6}$     & $86.1$    & $\tf{80.6}$    & $\tf{90.1}$   & $\tf{35.6}$    & $\tf{64.7}$   & $83.1$         & $\tf{82.6}$    & $\tf{82.6}$    & $\le 20$\\
\rowcolor{gray!15}Speedup  $\nearrow$           & $12.0\times$    & $12.1\times$   & $12.1\times$   & $11.0\times$  & $11.5\times$   & $11.9\times$  & $12.9\times$      & $11.9\times$   & $8.7\times$   & - \\\bottomrule
\end{tabular}}
\vspace{-0.5em}
\caption{{\ours} v.s. TinyBERT${_4}$~\cite{jiao2020tinybert} models with a $\sim$10$\times$ speedup. GD: general distillation, which distills the student model on a large unlabeled corpus. Train time is measured in GPU hours (see \autoref{sec:train_time} for details). The number of parameters for both models are around $5$M (around $95\%$ sparsity). {\ours} closes the gap between distillation and pruning with significantly less computation. Note that we remove data augmentation from TinyBERT for a fair comparison, see Table~\ref{tab:da} for experiments with augmented data.
}
\label{tab:tinybert}
\end{table*}

%!TEX root = ../main.tex

\begin{table}[t]
\centering
\resizebox{0.82\columnwidth}{!}{%
\begin{tabular}{lcc} \toprule
Task & \multicolumn{1}{c}{TinyBERT$_4$} & \multicolumn{1}{c}{{\ours} (ours)} \\
\midrule
SST-2 & $89.7\to91.6$ & $90.6\to\bf{92.4}$ \\
QNLI & $86.7\to\bf{87.6}$ & $86.1\to86.8$ \\
RTE & $63.2\to62.5$ & $64.7\to\bf{67.5}$ \\
MRPC & $81.4\to83.6$ & $82.6\to\bf{84.6}$ \\ \bottomrule
\end{tabular}
}
\caption{{\ours} v.s. TinyBERT${_4}$ trained with task-specific data augmentation introduced in \citet{jiao2020tinybert}. All models have around $5$M parameters ($95\%$ sparsity) and achieve similar speedups (11-12$\times$). The numbers before $\rightarrow$ are without data augmentation.}

\label{tab:da}
\end{table}

\paragraph{Baselines}
We compare {\ours} against several baselines: {DistillBERT${_6}$}~\cite{sanh2019distilbert}, {TinyBERT${_6}$} and {TinyBERT${_4}$}~\cite{jiao2020tinybert}, {DynaBERT}~\cite{hou2020dynabert}, and {Block Pruning}~\cite{lagunas2021block} (see \autoref{app:baseline} for details).
% We compare against several strong pruning and distillation models, including 1) \tf{DistillBERT${_6}$} \cite{sanh2019distilbert}; 2) \tf{TinyBERT${_6}$} and \tf{TinyBERT${_4}$} \cite{jiao2020tinybert} both include general distillation for pretraining and task-specific distillation; 3) \tf{DynaBERT} \cite{hou2020dynabert}: a method that provides dynamic-sized models by specifying width and depth; 4) \tf{Block Pruning} \cite{lagunas2021block}: a pruning method coupled with prediction-layer distillation. We choose their strongest approach ``Hybrid Filled'' as our baseline.
We also compare to other pruning methods such as FLOP~\cite{wang2020structured}, LayerDrop~\cite{fan2020reducing}, Movement Pruning~\cite{sanh2020movement} and distillation methods such as MobileBERT~\cite{sun2020mobilebert} and AutoTinyBERT~\cite{yin2021autotinybert} in \autoref{app:more_baseline}.\footnote{We show these results in \autoref{app:more_baseline} as they are not directly comparable to {\ours}.}

For TinyBERT and DynaBERT, the released models are trained with task-specific augmented data. For a fair comparison, we train these two models with the released code without data augmentation.\footnote{For TinyBERT, the augmented data is $20\times$ larger than the original data, making the  training process significantly slower.} For Block Pruning, we train models with their released checkpoints on GLUE tasks and use SQuAD results from the paper.

% \paragraph{Sparsity} In {{\ours}}, multiple masks control a single parameter but it does not affect the sparsity calculation. Sparsity is calculated as the percentage of the number of pruned parameters out of the full model size (excluding embeddings).
% \footnote{In the remainder of the paper, we use sparsity and pruning rate interchangeably.}

\paragraph{Speedup evaluation} Speedup rate is a primary measurement we use throughout the paper as the compression rate does not necessarily reflect the actual improvement in inference latency. \footnote{Models with the same compression rate could have considerably different speedups.} We use an unpruned BERT$_{\text{base}}$ as the baseline and evaluate all the models with the same hardware setup on a single NVIDIA V100 GPU to measure inference speedup. The input size is 128 for GLUE tasks and 384 for SQuAD, and we use a batch size of 128. Note that the results might be different from the original papers as the environment for each platform is different.

\subsection{Main Results}
\paragraph{Overall performance} In \autoref{fig:main}, we compare the accuracy of {{\ours}} models to other methods in terms of both inference speedup and model size. {{\ours}} delivers more accurate models than distillation and pruning baselines at every speedup level and model size. Block Pruning~\cite{lagunas2021block}, a recent work that shows strong performance against TinyBERT${_6}$, is unable to achieve comparable speedups as TinyBERT${_4}$. Instead, {{\ours}} has the option to prune both layers and heads \& intermediate units and can achieve a model with a comparable or higher performance compared to  TinyBERT${_4}$ and all the other models. Additionally, DynaBERT performs much worse speed-wise because it is restricted to remove at most half of the {\mha} and {\ffn} layers.

 % speedup comparable to TinyBERT${_4}$ and a comparable or even better performance across the board.

% , mainly because it does not incorporate the pruning option for layers
% With the same number of parameters, a model with fewer layers runs faster during inference time (See more analysis in \todo{}).

% \subsection{10$\times$ Faster Model}

\paragraph{Comparison with TinyBERT$_4$}
In \autoref{tab:tinybert}, we show that {{\ours}} produces models with over $10 \times$ inference speedup and achieves comparable or even better performance than TinyBERT${_4}$. General distillation (GD), which distills information from a large corpus, is essential for training distillation models, especially for small-sized datasets (e.g., TinyBERT$_4$ w/o GD performs poorly on CoLA, RTE and STS-B). While general distillation could take up to hundreds of GPU hours for training, {{\ours}} trains for a maximum of 20 hours on a task-specific dataset with a single GPU. We argue that pruning approaches---trained with distillation objectives like {{\ours}}---are more economical and efficient in achieving compressed models.

We further compare {\ours} with TinyBERT${_4}$ under the data augmentation setting in \autoref{tab:da}. As the augmented dataset is not publicly released, we follow its GitHub repository to create our own augmented data.
We train {\ours} with the same set of augmented data and find that it still outperforms TinyBERT$_4$ on most datasets.\footnote{We only conduct experiments with data augmentation on four datasets because training on augmented data is very expensive. For example, training on the augmented dataset for MNLI takes more than $200$ GPU hours in total. See more details in \autoref{app:tinybert4_reimplementation}.}

\subsection{Ablation Study}
\label{sec:ablation}

%!TEX root = ../main.tex

\begin{table*}[t]
\centering
\resizebox{2.0\columnwidth}{!}{%
\begin{tabular}{@{}lcccccccccccc@{}}
\toprule
                      & \multicolumn{2}{c}{QNLI ($60 \%$)} & \multicolumn{2}{c}{QNLI ($95 \%$)}& \multicolumn{2}{c}{MNLI ($60 \%$)} & \multicolumn{2}{c}{MNLI ($95 \%$)} & \multicolumn{2}{c}{SQuAD ($60 \%$)} & \multicolumn{2}{c}{SQuAD ($95 \%$)} \\
                      & \CC $\nearrow$ & acc & \CC $\nearrow$ & acc & \CC $\nearrow$ & acc & \CC $\nearrow$ & acc & \CC $\nearrow$ & F1 & \CC $\nearrow$ & F1 \\
                    \midrule

{\ours}            & \CC ${2.1}\times$          & $\tf{91.8}$          & \CC ${12.1}\times$         & $\tf{86.1}$   &\CC $2.1\times$ &	$85.1$	& \CC ${12.1}\times$	& $\tf{80.6}$      & \CC ${2.0}\times$           & $\tf{89.1}$             & \CC ${8.7}\times$          & $\tf{82.6}$              \\
\tableindent $-$hidden          & \CC ${2.2}\times$          & ${91.3}$          & \CC $13.3\times$         & $85.6$       & \CC $2.1\times$ &  $\tf{85.2}$              & \CC $13.7\times$              & $79.8$   & \CC $2.0\times$ & $88.7$ &  \CC $9.7\times$ &    $80.8$            \\
\tableindent $-$layer \& hidden & \CC ${2.2}\times$          & ${91.3}$          & \CC $7.2\times$          & $84.6$         & \CC $2.1\times$ & $84.8$ & \CC $7.0\times$ & $78.4$ &  \CC $2.1\times$           & $88.5$          & \CC $6.4\times$          & $74.1$          \\ \midrule
{\ours}            & \CC ${2.1}\times$          & ${91.8}$          & \CC $\tf{12.1}\times$         & ${86.1}$   & \CC $2.1\times$ &	$85.1$	& \CC $\tf{12.1}\times$	& $80.6$      & \CC ${2.0}\times$           & ${89.1}$             & \CC $\tf{8.7}\times$          & ${82.6}$              \\
\tableindent $-$layer & \CC $2.1\times$                    & $91.5$          & \CC ${8.3}\times$         & ${86.7}$       & \CC ${2.1}\times$ &    ${85.4}$            & \CC $8.4\times$              & 80.6   & \CC ${2.0}\times$ & ${89.1}$ &   \CC $7.9\times$ &    $80.5$  \\

\bottomrule
\end{tabular}}
\vspace{-0.5em}
\caption{Ablation studies on pruning units on QNLI, MNLI and SQuAD. $\nearrow$: speedup. The pruned models of a sparsity $60\%$ and $95\%$ have a model size of 34M and 5M respectively. $-$layer: When we do not prune entire layers (no ${z}_{\mathrm{MHA}}$ or ${z}_{\mathrm{FFN}}$), the speed-ups are greatly reduced for a high sparsity e.g., 95\%;  $-$hidden: when we remove the mask variables corresponding to hidden units ($\mathbf{z}_{\mathrm{hidn}}$), we observe a significant drop in accuracy. 
}
\label{tab:units}
\end{table*}

\paragraph{Pruning units} \label{sec:pruning_units} We first conduct an ablation study to investigate how additional pruning units such as {\mha} layers, {\ffn} layers and hidden units in {{\ours}} affect model performance and inference speedup beyond the standard practice of pruning heads and {\ffn} dimensions. We show results in \autoref{tab:units} for models of similar sizes. Removing the option to prune hidden dimensions ($\mathbf{z}_{\mathrm{hidn}}$) leads to a slightly faster model with a performance drop across the board and we find that it removes more layers than {{\ours}} and does not lead to optimal performance under a specific sparsity constraint.  In addition, removing the layer masks (${\mathbf{z}}_{\mathrm{MHA}}$, ${\mathbf{z}}_{\mathrm{FFN}}$) brings a significant drop in speedup on highly compressed models (95\%, 5M). This result shows that even with the same amount of parameters, different configurations for a model could lead to drastically different speedups. However, it does not affect the lower sparsity regime (60\%, 34M). In short, by placing masking variables at different levels, the optimization procedure is incentivized to prune units accordingly under the sparsity constraint while maximizing the model performance.

%!TEX root = ../main.tex

% \begin{table}[t]
% \resizebox{1.0\columnwidth}{!}{%
% \begin{tabular}{@{}lccccc@{}} \toprule
%                     & SST-2 & QNLI  & MNLI  & QQP & SQuAD \\ \midrule
% Fixed Hidn Distil. & $90.02$ & $85.78$ & $80.54$ & $89.98$ & $80.93$ \\ \midrule
% {\ours}        & $90.60$ & $\bf{86.14}$ & $\bf{80.55}$ &  $\bf{90.12}$ & X    \\
% -  Hidn Distil.     & $\bf{91.06}$ & $85.08$ & $79.66$ & $89.76$ &  $82.52$ \\
% No Distil.          & $86.58$ & $84.15$ & $78.15$ & $88.13$ & $75.78$ \\ \bottomrule
% \end{tabular}}
% \caption{Ablation study over distillation objectives on pruned models with a $95\%$ sparsity. -Hidn Distil. denotes removing hidden layer distillation.}
% \label{tab:distillation}
% \end{table}

\begin{table}[t]
\resizebox{1.0\columnwidth}{!}{%
\begin{tabular}{@{}lccccc@{}} \toprule
                    & SST-2 & QNLI  & MNLI   & SQuAD \\ \midrule
{\ours}        & $90.6$ & $\bf{86.1}$ & $\bf{80.6}$ &   \bf{82.6}    \\
\tableindent $-\mathcal{L}_{\mathtt{layer}}$     & $\bf{91.1}$ & $85.1$ & $79.7$ &  $82.5$ \\
\tableindent $-\mathcal{L}_{\mathtt{pred}}, \mathcal{L}_{\mathtt{layer}}$         & $86.6$ & $84.2$ & $78.2$ & $75.8$ \\
\midrule
Fixed Hidn Distil. & $90.0$ & $85.8$ & $80.5$ &  $80.9$ \\
\bottomrule
\end{tabular}}
\vspace{-0.5em}
\caption{Ablation study of different distillation objectives on pruned models with sparsity $=95\%$. Fixed hidden distillation: simply matching each layer of the student and the teacher model, see \S\ref{sec:ablation} for more details. In \S\ref{app:layerdistill}, we show that the dynamic layer distillation objective improves model performance more significantly on lower sparsity rates. }
\label{tab:distillation}
\end{table}

% \begin{table}[t]
% \resizebox{1.0\columnwidth}{!}{%
% \begin{tabular}{@{}lccccc@{}} \toprule
%                     & SST-2 & QNLI  & MNLI   & SQuAD & Avg.\\ \midrule

% No distillation        & $86.6$ & $84.2$ & $78.2$ & $75.8$ & $81.2$ \\
% \tableindent $+\mathcal{L}_{\mathtt{pred}}$     & $\bf{91.1}$ & $85.1$ & $79.7$ &  $82.5$  & $84.6$ \\
% \tableindent $+ \mathcal{L}_{\mathtt{naive\,\,layer}}, \mathcal{L}_{\mathtt{pred}}$ & $90.0$ & $85.8$ & $80.5$ & $80.9$ & $84.3$\\ \midrule
% \tableindent{$+ \mathcal{L}_{\mathtt{layer}}, \mathcal{L}_{\mathtt{pred}}$}       & $90.6$ & $\bf{86.1}$ & $\bf{80.6}$ &   \bf{82.6}  &  $\bf{85.0}$  \\
% \bottomrule
% \end{tabular}}
% \vspace{-0.5em}
% \caption{Ablation study of different distillation objectives on pruned models with sparsity $=95\%$. Fixed hidden distillation: simply matching each layer of the student and the teacher model, see \S\ref{sec:ablation} for more details. In \S\ref{app:layerdistill}, we show that the dynamic layer distillation objective improves model performance more significantly on lower sparsity rates. }
% \label{tab:distillation}
% \end{table}

\paragraph{Distillation objectives} We also ablate on distillation objectives to see how each part contributes to the performance of {{\ours}} in \autoref{tab:distillation}. We first observe that removing distillation entirely leads to a performance drop up to 1.9-6.8 points across various datasets, showing the necessity to combine pruning and distillation for maintaining performance. The proposed hidden layer distillation objective dynamically matches the layers from the teacher model to the student model. We also experiment with a simple alternative i.e., ``Fixed Hidden Distillation'', which matches each layer from the teacher model to the corresponding layer in the student model -- if a layer is already pruned, the distillation objective will \ti{not} be added. We find that fixed hidden distillation underperforms the dynamic layer matching objective used for {{\ours}}.
Interestingly, the proposed dynamic layer matching objective consistently converges to a specific alignment between the layers of the teacher model and student model. For example, we find that on QNLI the training process dynamically matches the 3, 6, 9, 12 layers in the teacher model to 1, 2, 4, 9 layers in the student model.\footnote{Please refer to \S\ref{app:alignment} for more details.}  Moreover, as shown in the table, removing it hurts the performance for all the datasets except SST-2.

\begin{figure}[t]
\centering
\includegraphics[width=1.0\linewidth]{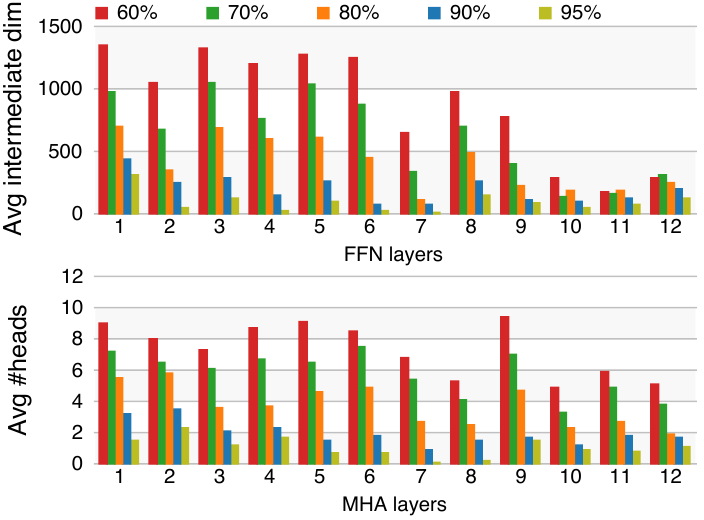}
\vspace{-1em}
\caption{The average intermediate dimensions at each FFN layer and the average number of heads at each MHA layer in the pruned models across five datasets (SST-2, MNLI, QQP, QNLI, and SQuAD 1.1). We study different sparsities $\{60\%, 70\%, 80\%, 90\%, 95\% \}$.}
\label{fig:arch_stats}
\end{figure}

%!TEX root = ../main.tex

\begin{table}[t]
\centering
\resizebox{0.9\columnwidth}{!}{%
\begin{tabular}{@{}ll@{}}
\toprule
{Dataset} & {Pruned Models} \\
\midrule
\multirow{3}{*}{SST-2}
& \texttt{\fst{mfmgmgngmgngngnfmfnfmfmf}} \\
& \texttt{\fst{mfngngngmgngngnfmfmgmfmf}} \\
& \texttt{\fst{mgngngngngnfmgnfmfmgmgmg}} \\
\midrule
\multirow{3}{*}{QNLI}
& \texttt{\fst{mfmgmfmgngngnfmgnfmgngng}} \\
& \texttt{\fst{mfmgmfmgngngngngmgngngmg}} \\
& \texttt{\fst{mfmgngmgngngngnfngmgnfmf}} \\

\midrule
\multirow{3}{*}{MNLI}
& \texttt{\fst{nfmgngnfmgngngngngngmfnf}} \\
& \texttt{\fst{mfmfmgngngngngngmfmgmgng}} \\
& \texttt{\fst{mfmgnfmgnfmgngngngmgngmg}} \\

\midrule
\multirow{3}{*}{QQP}
& \texttt{\fst{nfmgmgmgnfmgngngngnfnfmf}} \\
& \texttt{\fst{nfmgnfmgnfmgngnfmgngngnf}} \\
& \texttt{\fst{nfmgmfmgmfmgngnfmgngngmg}} \\

\midrule
\multirow{3}{*}{SQuAD}
& \texttt{\fst{nfmfmfmgngmgngnfmgngngnf}} \\
& \texttt{\fst{nfmgmfmgngngngnfmgnfmgnf}} \\
& \texttt{\fst{nfmfmgmgnfngngmfmgngngnf}} \\
    \bottomrule
\end{tabular}}
\vspace{-0.5em}
\caption{Remaining layers in the models pruned by {{\ours}} on different datasets. All models are pruned at a sparsity of $95\%$.
For each setting, we run the experiments three times to obtain three different pruned models.
\texttt{\fst{m}} represents a remaining MHA layer and \texttt{\fst{f}} represents a remaining FFN layer.}
\label{tab:layer_patterns}
\end{table}

\subsection{Structures of Pruned Models}
Finally, we study the pruned structures produced by {{\ours}}. We characterize the pruned models of sparsities $\{60\%, 70\%, 80\%, 90\%, 95\% \}$ on five datasets. For each setting, we run {{\ours}} three times.
\autoref{fig:arch_stats} demonstrates the number of remaining heads and intermediate dimensions of the pruned models for different sparsities.\footnote{We show more layer analysis in \autoref{app:layer_analysis}.}
Interestingly, we discover common structural patterns in the pruned models:
(1)~Feed-forward layers are significantly pruned across all sparsities. For example, at the $60\%$ sparsity level, the average number of intermediate dimensions in FFN layers after pruning is reduced by $71\%$ ($3,072 \to 884$), and the average number of heads in MHA is reduced by $39\%$ ($12 \to 7.3$). This suggests {\ffn} layers are more redundant than {\mha} layers.
(2) {{\ours}} tends to prune submodules more from upper layers than lower layers. For example, upper {\mha} layers have fewer remaining heads than lower layers on average.

Furthermore, we study the number of remaining {\ffn} and {\mha} layers and visualize the results in \autoref{tab:layer_patterns} for highly compressed models (sparsity $=95\%$). Although all the models are roughly of the same size, they present different patterns across datasets, which suggests that there exist different optimal sub-networks for each dataset. We find that on SST-2 and QNLI, the first {\mha} layer is preserved but can be removed on QQP and SQuAD. We also observe that some layers are particularly important across all datasets. For example, the first {\mha} layer and the second {\mha} layer are preserved most of the time, while the middle layers are often removed.
Generally, the pruned models contain more {\mha} layers than {\ffn} layers (see \autoref{app:layer_analysis}), which suggests that {\mha} layers are more important for solving downstream tasks. Similar to \citet{press2019improving}, we find that although standard Transformer networks have interleaving {\ffn} layers and {\mha} layers, in our pruned models, adjacent {\ffn}/{\mha} layers could possibly lead to a better performance.

%!TEX root = ../main.tex

\section{Related Work}

Structured pruning has been widely explored in computer vision, where channel pruning \cite{he2017channel, luo2017thinet, liu2017learning, liu2018rethinking, liu2019metapruning, molchanov2019importance, Guo_2020_CVPR} is a standard approach for convolution neural networks. The techniques can be adapted to Transformer-based models as introduced in \S\ref{subsec:bgpruning}. 
Unstructured pruning is another major research direction, especially gaining popularity in the theory of Lottery Ticket Hypothesis~\cite{frankle2018the,zhou2019deconstructing, Renda2020Comparing, frankle2020linear, chen2020lottery}. Unstructured pruning produces models with high sparsities \cite{sanh2020movement, xu2021rethinking, huang2021sparse} yet hardly bring actual inference speedups. Developing computing platforms for efficient sparse tensor operations is an active research area. DeepSparse\footnote{https://github.com/neuralmagic/deepsparse} is a CPU inference engine that leverages unstructured sparsity for speedup. \citet{huang2021sparse} measure the real inference speedup induced by unstructured pruning on Moffett AI’s latest hardware platform ANTOM. We do not directly compare against these methods because the evaluation environments are different.

While all the aforementioned methods produce task-specific models through pruning, several works explore upstream pruning where they prune a large pre-trained model with the masked language modeling task. \citet{chen2020lottery} show a 70\%-sparsity model retains the MLM accuracy produced by iterative magnitude pruning. \citet{zafrir2021prune} show the potential advantage of upstream unstructured pruning against downstream pruning.  We consider applying {\ours} for upstream pruning as a promising future direction to produce task-agnostic models with flexible structures.

Besides pruning, many other techniques have been explored to gain inference speedups for Transformer models, including distillation as introduced in \S\ref{subsec:dis}, quantization~\cite{shen2020q,fan2021training}, dynamic inference acceleration~\cite{xin2020deebert} and matrix decomposition~\cite{noach2020compressing}. We refer the readers to \newcite{ganesh2021compressing} for a comprehensive survey.

%!TEX root = ../main.tex

\section{Conclusion}
We propose {\ours}, a structured pruning approach that incorporates all levels of pruning, including MHA/FFN layers, individual heads, and hidden dimensions for Transformer-based models. Coupled with a distillation objective tailored to structured pruning, we show that {\ours} compresses models into a rather different structure from standard distillation models but still achieves competitive results with more than $10\times$ speedup. We conclude that task-specific structured pruning from large-sized models could be an appealing replacement for distillation to achieve extreme model compression, without resorting to expensive pre-training or data augmentation. Though {\ours} can be directly applied to structured pruning for task-agnostic models, we frame the scope of this work to task-specific pruning due to the complexity of the design choices for upstream pruning. We hope that future research continues this line of work, given that pruning from a large pre-trained model could possibly incur less computation compared to general distillation and leads to more flexible model structures.

\section*{Acknowledgements}

% \danqi{Acknowledgements sections are usually not numbered?}

The authors thank  Tao Lei from Google Research, Ameet Deshpande, Dan Friedman, Sadhika Malladi from Princeton University and the anonymous reviewers for their valuable feedback on our paper. This research is supported by a Hisashi and Masae Kobayashi *67 Fellowship and a Google Research Scholar Award.

\bibliography{ref}
\bibliographystyle{acl_natbib}

\clearpage
\appendix

%!TEX root = ../main.tex

\label{sec:appendix}

\section{Reproducibility \& Hyperparameters}
\label{app:hyperparameters}
We report the hyperparameters that we use in our experiments in \autoref{tab:hyperpara}.
\begin{table}[h]
\centering
\resizebox{1.0\columnwidth}{!}{%
\begin{tabular}{@{}lc@{}}
\toprule
{Hyperparameter} &   \\
\midrule
$\lambda$ & {$0.1, 0.3, 0.5$}\\
distillation temperature $t$ & {$2$} \\
finetuning epochs & $20$ \\
finetuning learning rate & $1e-5, 2e-5, 3e-5$ \\
training learning rate & $2e-5$ (GLUE), $3e-5$ (SQuAD)\\
batch size & $32$ (GLUE), $ 16$ (SQuAD)\\ \bottomrule
\end{tabular}
}
\caption{Hyperparemeters in the experiments.}
\label{tab:hyperpara}
\end{table}

For four relatively larger GLUE datasets, MNLI, QNLI, SST-2 and QQP, and SQuAD, we train the model for 20 epochs in total and finetune the finalized sub-network for another 20 epochs. In the first 20 epochs, following \citet{lagunas2021block} and \citet{wang2020structured}, we first finetune the model with the distillation objective for 1 epoch, and then start pruning with a linear schedule to achieve the target sparsity within 2 epochs. For the four small GLUE datasets, we train the model for 100 epochs in total and finetune for 20 epochs. We finetune the model with the distillation objective for 4 epochs and prune till the target sparsity within the next 20 epochs. Note that even if the final sparsity is achieved, the pruning process keeps searching better performing structures in the rest of the training epochs. In addition, we find that finetuning the final subnetwork is essential for high sparsity models. Hyperparameters like $\lambda$, batch size, and learning rate do not generally affect performance much.

\section{Optimization Details}
\label{app:optimization}
\citet{louizos2018learning} propose $l_0$ optimization for model compression where the masks are modeled with hard concrete distributions as follows:

\begin{gather}
\begin{aligned}
    \mathbf{u} & \sim U(0, 1) \\
    % \mathbf{s} & = \text{sigmoid}((\log \mathbf{u} - \log (1-\mathbf{u}) + \log\alpha) / \beta) \\
    \mathbf{s} & = \text{sigmoid}\left(\frac{1}{\beta}{\log \mathbf{\frac{u}{1-u}} + \log\alpha}\right) \\
    \mathbf{ \tilde{s}} & = \mathbf{s} \times (r-l) + l \\
    \mathbf{z} & = \min(1, \max(0,  \mathbf{\tilde{s}})). \label{eq:hard_concrete} \\ \nonumber
\end{aligned}
\end{gather}
${U(0, 1)}$ is a uniform distribution in the interval $[0, 1]$; $l < 0$ and $r > 0$ are two constants that stretch the sigmoid output into the interval $(l, r)$. $\beta$ is a hyperparameter that controls the steepness of the sigmoid function and $\log \alpha$ are the main learnable parameters. We learn the masks through updating the learnable parameters of the distributions from which the masks are sampled in the forward pass.

In our preliminary experiments, we find that optimizing $\lambda \|\mf{z}\|_0$ with different learning rates and pruning schedules may converge to models of drastically different sizes. Hence, we follow \newcite{wang2020structured} to add a Lagrangian term, which imposes an equality constraint $\hat s = t$ by introducing a violation penalty:

\begin{gather}
    \mathcal{L}_c = \lambda_1 \cdot (\hat s - t) + \lambda_2 \cdot (\hat s - t)^2, \nonumber
\end{gather}

where ${\hat s}$ is the expected model sparsity calculated from $\mathbf{z}$ and $t$ is the target sparsity.

\section{Details of Baseline Methods}
\label{app:baseline}
We compare against several strong pruning and distillation models, including 1) \tf{DistillBERT${_6}$} \cite{sanh2019distilbert}; 2) \tf{TinyBERT${_6}$} and \tf{TinyBERT${_4}$} \cite{jiao2020tinybert} both include general distillation for pretraining and task-specific distillation; 3) \tf{DynaBERT} \cite{hou2020dynabert}: a method that provides dynamic-sized models by specifying width and depth; 4) \tf{Block Pruning} \cite{lagunas2021block}: a pruning method coupled with prediction-layer distillation. We choose their strongest approach ``Hybrid Filled'' as our baseline. 

\section{Data Statistics}
\label{app:data}
We show train sizes and metrics for each dataset we use in \autoref{tab:data}.

\begin{table}[h]
\centering
\resizebox{0.75\columnwidth}{!}{%
\begin{tabular}{lrr} \toprule
Task  & Train Size & Metric \\ \midrule
SST-2 & $67$k  & accuracy      \\
QNLI  & $105$k & accuracy       \\
MNLI  & $393$k & accuracy      \\
QQP   & $364$k & accuracy \\
CoLA  & $8.5$k & Matthews corr.\\
RTE   & $2.5$k & accuracy         \\
STS-B & $7$k & Spearman corr. \\
MRPC  & $3.7$k  & accuracy        \\
SQuAD & $88$k   & F1     \\ \bottomrule
\end{tabular}
}
\caption{Data statistics of GLUE and SQuAD datasets.}
\label{tab:data}
\end{table}

\section{TinyBERT$_{4}$ w/ Data Augmentation}
\label{app:tinybert4_reimplementation}
We conduct task-specific distillation with the script provided by the TinyBERT repository.\footnote{https://github.com/huawei-noah/Pretrained-Language-Model/tree/master/TinyBERT} However, our reproduced results are slightly lower than the reported results in \citep{jiao2020tinybert}. The difference between these two sets of
scores may stem from augmented data or teacher models.  Note that the authors of TinyBERT did not release the augmented dataset. We run their codes to obtain augmented datasets. We compare CoFi and TinyBERT under the same setting where we use the same teacher model and the same set of augmented data.
\begin{table}[h]
\resizebox{1.0\columnwidth}{!}{%
\begin{tabular}{lcccc} \toprule
            & SST-2 & QNLI & RTE  & MRPC \\ \midrule
TinyBERT$_4$ reimpl. & 91.6  & 87.6 & 62.5 & 83.6 \\
\citet{jiao2020tinybert}            & 92.7  & 88.0 & 65.7 & 85.7 \\ \bottomrule
\end{tabular}}
\caption{Re-implemented (TinyBERT${_4}$ reimpl.) results and the results reported in \citet{jiao2020tinybert}.}
\end{table}

\section{Additional Comparisons}
\label{app:more_baseline}

\subsection{Comparison to Movement Pruning}
\begin{figure}[h]
\centering
\includegraphics[width=1\linewidth]{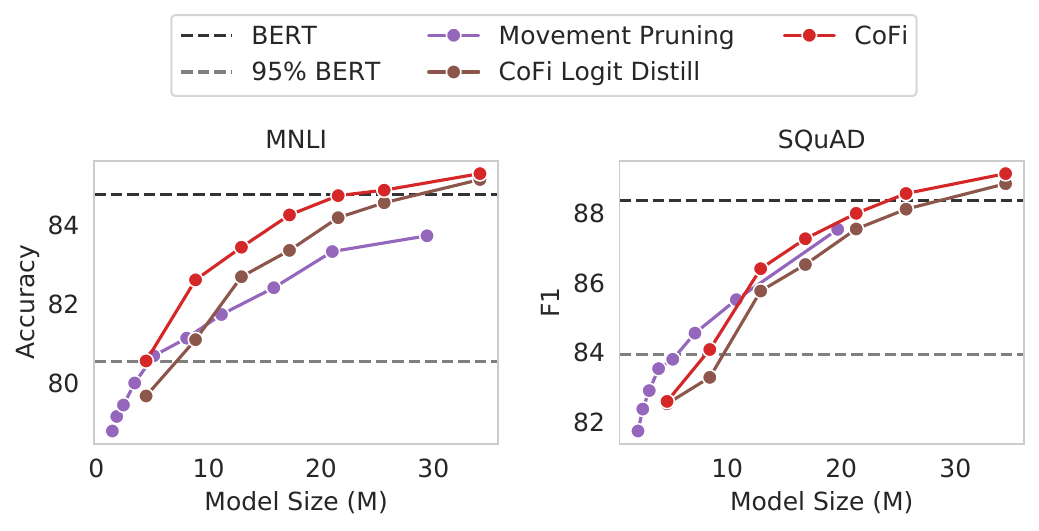}
\caption{{\ours} v.s. Movement Pruning (unstructured pruning). {\ours} Logit Distill denotes that we run {\ours} with prediction-layer distillation only as Movement Pruning.}
\label{fig:unstructured_pruning}
\end{figure}

We compare {\ours} with a state-of-the-art \tf{unstructured pruning} method, Movement Pruning \cite{sanh2020movement} in \autoref{fig:unstructured_pruning}. As Movement Pruning is trained with prediction-layer (logit) distillation only, we also show results of {\ours} trained with the same distillation objective. We observe that {\ours} largely outperforms Movement Pruning even without layerwise distillation on MNLI and is comparable to SQuAD on models with a size over $10M$ parameters. {\ours}, as a structured pruning method, is less performant on models of a sparsity up to $95\%$, as pruning flexibility is largely restricted by the smallest pruning unit. However, pruned models from {\ours} achieve $2-11\times$ inference speedups while no speedup gains are achieved from Movement Pruning.

\subsection{Comparison to Block Pruning}

\begin{figure*}[t]
\centering
\includegraphics[width=1\linewidth]{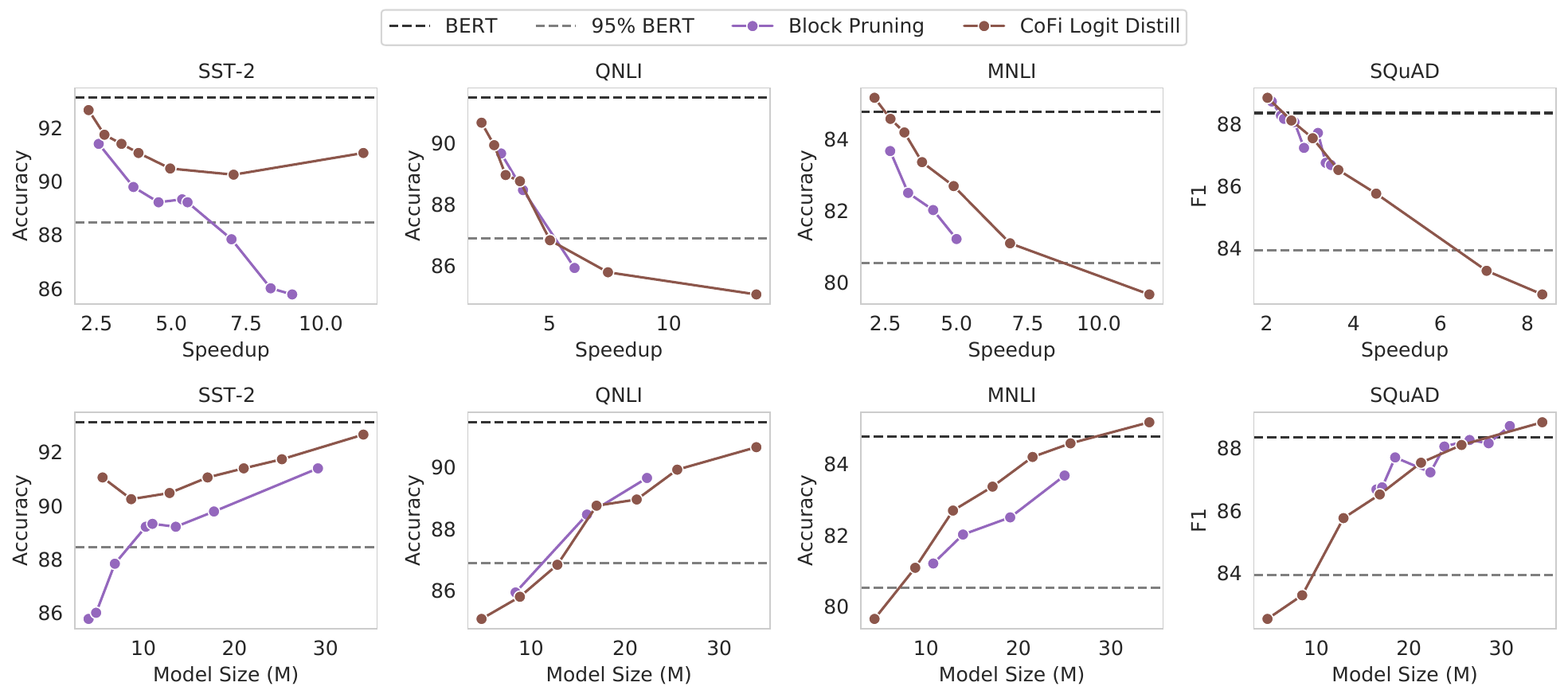}
\caption{{\ours} v.s. Block Pruning with the same distillation objective -- prediction-layer distillation (Logit Distill). {\ours} still outperforms or is on par with Block Pruning.}
\label{fig:block_pruning}
\end{figure*}

In \autoref{fig:block_pruning}, we compare {\ours} with Block Pruning while unifying the distillation objective. Without the layer distillation objective, {\ours} still outperforms or is on par with Block Pruning. Block Pruning never achieves a speedup of 10 even the pruned model is of a similar size as {\ours} (SST-2), backing up our argument that pruning layers for high sparsity models is the key to high speedups.

\subsection{More Baselines}
We show additional pruning and distillation methods that are not directly comparable to {\ours} in \autoref{tab:more_on_sp_distill}. {\ours} still largely outperforms these baselines even though these methods hold an inherent advantage due to a stronger teacher or base model. % \danqi{I feel you need to say a bit about your model sizes or pruning rate, given they are far from the $12\times$ models.}
\begin{table}[h]
\resizebox{1.0\columnwidth}{!}{%
\begin{tabular}{@{}lccccc@{}} \toprule
          & $\nearrow$ & SST-2 & QNLI & MNLI & SQuAD \\ \midrule
% \citet{sanh2020movement}$^{\diamond}$ & 1.0$\times$ & - & - & 81.8 & 84.9 \\
\citet{wang2020structured}$^{\ddagger}$            & 1.5$\times$    & 92.1  & 89.1 & -    & 85.4  \\
\citet{sajjad2020poor} & 2.0$\times$    & 90.3  & -    & 81.1 & -     \\
\citet{fan2020reducing}$^{\ddagger}$       & 2.0$\times$    & 93.2  & 89.5 & 84.1 & -     \\
\citet{sun2020mobilebert}$^{\diamond}$      & 2.3$\times$    & 92.1  & 91.0   & 83.9 & 90.3  \\
\citet{yin2021autotinybert}$^{\spadesuit}$ & 4.3$\times$ & 91.4 & 89.7 & 82.3 & 87.6 \\
\midrule
{\ours} (ours)    & 2.0$\times$    & 93.0    & 91.8 & 85.3 & 89.1  \\
{\ours} (ours)    & 4.6$\times$    & 92.6    & 89.7 & 83.4 & 86.4  \\\bottomrule
\end{tabular}}
\caption{More pruning and distillation baselines.  $\nearrow$: speedup. $\ddagger$ denotes that the model prunes from a RoBERTa${_{\text{base}}}$ model. $\spadesuit$: AutoTinyBERT is distilled from an ELECTRA${_{\text{base}}}$ model. $^{\diamond}$: MobileBERT \cite{sun2020mobilebert} has specialized architecture designs and trains their own teacher model from the scratch. {\ours} models have a model size of $34$M and $13$M respectively, corresponding to a $60\%$ and $85\%$ sparsity.}
\label{tab:more_on_sp_distill}
\end{table}

% $\dagger$ denotes that it is a pruning method without distillation training.

\section{More Analyses on Layer Distillation}

\subsection{Layer Alignment}
\label{app:alignment}
We find that the alignment between the layers of the student model and the teacher model shifts during the course of training. To take SST-2 for an example, as the training goes on, the model learns the alignment to match the $7, 9, 10, 11$ layers of the student model to the $3, 6, 9, 12$ layers of the teacher model. For QQP, the model eventually learns to map $2, 5, 8, 11$ layers to the four layers of the teacher model. The final alignment shows that our dynamic layer matching distillation objective can find task-specific alignment and improve performance.

\subsection{Ablation on Distillation Objectives}
\label{app:layerdistill}
In \autoref{tab:ablation_layer_more}, we show ablation studies on adding the dynamic layer distillation onto prediction distillation across all sparsities.
Using the layer distillation loss clearly helps improve the performance on all sparsity rates and different tasks.

\section{FFN/MHA Layers in Pruned Models}
\label{app:layer_analysis}
\begin{figure}[h!]
\centering
\includegraphics[width=1\linewidth]{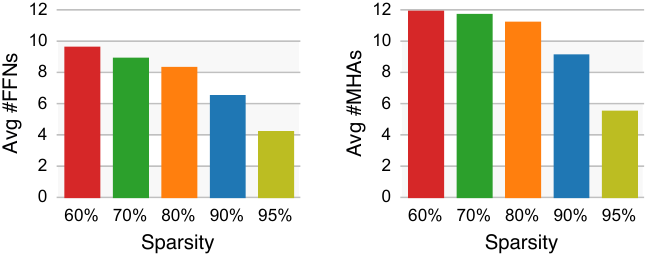}
\caption{The average number of FFN layers and MHA layers in the pruned models at different sparsities.
% \danqi{If you have time, consider replacing the x-axis from pruning rate to sparsity.}
}
\label{fig:avg_layers}
\end{figure}

\autoref{fig:avg_layers} shows the average number of FFN layers and MHA layers in the pruned models by {\ours}. We study different sparsities $\{60\%, 70\%, 80\%, 90\%, 95\%\}$. It is clear that when the sparsity increases, the pruned models become shallower (i.e., the number of layers becomes fewer). Furthermore, we find that the pruned models usually have more MHA layers than FFN layers. This may indicate that MHA layers are more important for solving these downstream tasks than FFN layers.

% \danqi{Say something about how expensive these experiments are and why we don't use data augmentation in our main experiments.}
% \danqi{Can you add the numbers without DA to this table too? It is great to see the difference. Can make it double column if needed. I think this table is important and shouldn't be buried in the middle.}

\begin{table*}[t]
\centering
\resizebox{2.0\columnwidth}{!}{
\begin{tabular}{lcccccccc}
\toprule
         & \multicolumn{2}{c}{SST-2}    & \multicolumn{2}{c}{QNLI}     & \multicolumn{2}{c}{MNLI}     & \multicolumn{2}{c}{SQuAD}    \\
sparsity & $\mathcal{L}_{\texttt{pred}}$ &  $+\mathcal{L}_{\texttt{layer}}$&  $\mathcal{L}_{\texttt{pred}}$ & $+\mathcal{L}_{\texttt{layer}}$ & $\mathcal{L}_{\texttt{pred}}$ & $+\mathcal{L}_{\texttt{layer}}$ & $\mathcal{L}_{\texttt{pred}}$ & $+\mathcal{L}_{\texttt{layer}}$ \\ \midrule
$60\%$      & $92.66$   & $93.00~(+0.34)$              & $90.66$   & $91.84~(+1.18)$              & $85.16$   & $85.31~(+0.15)$              & $88.84$   & $89.13~(+0.29)$              \\
$70\%$       & $91.74$   & $93.00~(+1.26)$              & $89.93$   & $91.29~(+1.36)$              & $84.57$   & $84.89~(+0.32)$              & $88.11$   & $88.56~(+0.45)$              \\
$75\%$      & $91.40$   & $92.89~(+1.49)$              & $88.96$   & $91.31~(+2.35)$              & $84.19$   & $84.75~(+0.56)$              & $87.54$   & $87.99~(+0.45)$              \\
$80\%$       & $91.06$   & $92.89~(+1.83)$              & $88.76$   & $90.43~(+0.67)$              & $83.36$   & $84.26~(+0.90)$              & $86.52$   & $87.26~(+0.74)$              \\ 
$85\%$      & $90.48$   & $92.55~(+2.07)$              & $86.84$   & $89.69~(+2.85)$ & $82.69$   & $83.44~(+0.75)$              & $85.76$   & $86.40~(+0.64)$              \\
$90\%$       & $90.25$   & $91.51~(+1.26)$              & $85.80$   & $88.89~(+3.19)$              & $81.09$   & $82.61~(+1.52)$              & $83.28$   & $84.08~(+0.80)$              \\
$95\%$      & $91.06$   & $90.37~(-0.69)$              & $85.08$   & $86.14~(+1.06)$              & $79.66$   & $80.55~(+0.89)$              & $82.52$   & $82.59~(+0.07)$    \\ 
% \midrule
% average & $91.24$ &	$92.32$	& $88.01$ &	$89.94$ &	$82.96$ &	$83.69$ &	$86.08$ &	$86.57$ \\
\bottomrule
\end{tabular}
}
\caption{Ablation study on the proposed layer distillation objective across all sparsities. 
% The layer distillation objective leads to average gains up to $0.5-1.9$ points. \danqi{What is the point of averaging across sparsities, not datasets??} \danqi{If the goal is to show the delta for different sparisty level, consider adding $(+0.34)$ next to $93.00$.}
}
\label{tab:ablation_layer_more}
\end{table*}

\section{RoBERTa Pruning}
\label{app:roberta}
\begin{figure*}[h!]
\centering
\includegraphics[width=0.7\linewidth]{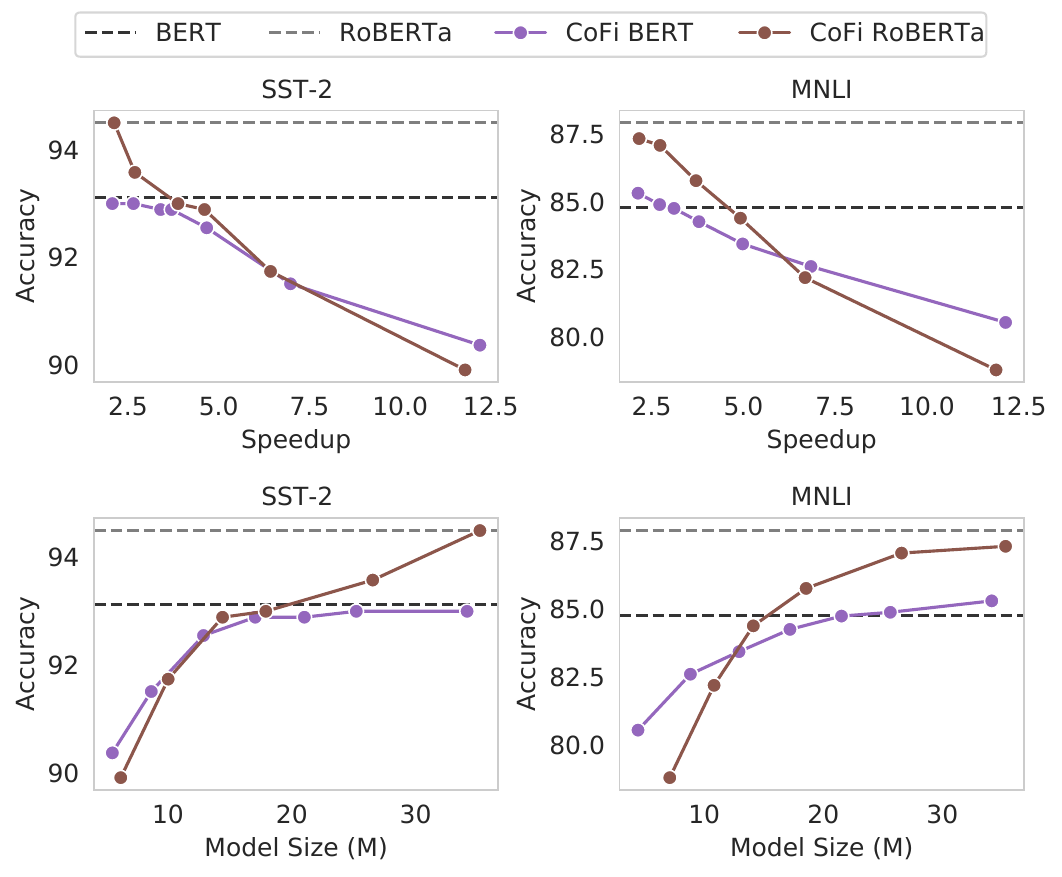}
\caption{{\ours} with BERT and RoBERTa on SST-2 and MNLI.}
\label{fig:roberta}
\end{figure*}

\noindent We show {\ours} results with RoBERTa in \autoref{fig:roberta} across sparsities from $60\%$ to $95\%$. Similar to BERT, models with $60\%$ weights pruned are able to maintain the performance of a full model. Pruning from RoBERTa outperforms BERT on sparsities lower than $90\%$ but as the sparsity further increases, BERT surpasses RoBERTa. Similar patterns are observed from DynaBERT \cite{hou2020dynabert}.

\section{Training Time Measurement}
\label{sec:train_time}
We use NVIDIA RTX 2080Ti GPUs to measure the training time of TinyBERT.
For the general distillation step of TinyBERT, we measure the training time on a small corpus (containing 10.6M tokens) on 4 GPUs and estimate the training time on the original corpus (containing 2500M tokens) by scaling the time with the corpus size difference. Specifically, it takes $430$s to finish one epoch on 10.6M tokens with 4 GPUs, and we estimate that it will take 338 GPU hours (or 3.5 days with 4 GPUs) to finish three epochs on 2500M tokens.

% \begin{table}[h]
% \begin{tabular}{@{}lcc@{}}
% \toprule
%               & general distill. & task distill. \\ \midrule
% TinyBERT$_4$   & $\sim 338$             & $\le10$      \\
% {\ours} & 0                & $\le20$      \\ \bottomrule
% \end{tabular}
% \caption{Computational time (in GPU hours) of distillation v.s. pruning.}
% \label{tab:train_time}
% \end{table}

% \section{EarlyBERT Comparison}
% \begin{table}[]
% \centering
% \resizebox{1.0\columnwidth}{!}{%
% \begin{tabular}{lccccc} \toprule
%              & MNLI  & QNLI  & QQP   & SST-2 & SQuAD \\ \midrule
% EarlyBERT    & 81.81 & 89.18 & 90.06 & 90.71 & 86.13 \\
% {\ours} & 83.34 & 90.06 & 90.97 & 91.86 & 87.19 \\ \bottomrule
% \end{tabular}}
% \caption{{\ours} v.s. EarlyBERT without distillation. The two methods achieve a $2 \times$ speedup while }
% \label{tab:earlybert}
% \end{table}

\end{document}